\documentclass[conference]{IEEEtran}
\IEEEoverridecommandlockouts
\usepackage{cite}
\usepackage{amsmath,amssymb,amsfonts}
\usepackage{algorithmic}
\usepackage{graphicx}
\usepackage{textcomp}
\usepackage{xcolor}

\usepackage{array}
\usepackage[caption=false,font=small,labelfont=sf,textfont=sf]{subfig}
\usepackage{stfloats}
\usepackage{url}
\usepackage{verbatim}
\usepackage{booktabs}
\usepackage{multirow}
\usepackage[lined,boxed,commentsnumbered,ruled]{algorithm2e}
\usepackage{siunitx}
\usepackage{amsthm}

\def\BibTeX{{\rm B\kern-.05em{\sc i\kern-.025em b}\kern-.08em
    T\kern-.1667em\lower.7ex\hbox{E}\kern-.125emX}}
\begin{document}

\title{Large Language Models as Evolutionary Optimizers}





\author{\IEEEauthorblockN{Shengcai Liu}
\IEEEauthorblockA{\textit{Centre for Frontier AI Research, A*STAR}\\
}
\IEEEauthorblockA{\textit{Department of Computer Science and Engineering}\\
\textit{Southern University of Science and Technology}\\
liu\_shengcai@cfar.a-star.edu.sg}
\and
\IEEEauthorblockN{Caishun Chen}
\IEEEauthorblockA{\textit{Centre for Frontier AI Research}\\
\textit{A*STAR}\\
chen\_caishun@cfar.a-star.edu.sg}
\and
\IEEEauthorblockN{Xinghua Qu}
\IEEEauthorblockA{\textit{Tianqiao \& Chrissy Chen Institute}\\
quxinghua17@gmail.com}
\and
\IEEEauthorblockN{Ke Tang}
\IEEEauthorblockA{\textit{Guangdong Provincial Key Laboratory of}\\
\textit{Brain-Inspired Intelligent Computation}\\
\textit{Southern University of Science and Technology}\\
tangk3@sustech.edu.cn}
\and
\IEEEauthorblockN{Yew-Soon Ong}
\IEEEauthorblockA{\textit{Centre for Frontier AI Research, A*STAR}\\
\textit{School of Computer Science and Engineering, Data Science and}\\
\textit{Artificial Intelligence Research Centre, Nanyang Technological University}\\
asysong@ntu.edu.sg}
\thanks{Correspondence to asysong@ntu.edu.sg}}

\maketitle

\begin{abstract}
Evolutionary algorithms (EAs) have achieved remarkable success in tackling complex combinatorial optimization problems.
However, EAs often demand carefully-designed operators with the aid of domain expertise to achieve satisfactory performance.
In this work, we present the first study on large language models (LLMs) as evolutionary combinatorial optimizers.
The main advantage is that it requires minimal domain knowledge and human efforts, as well as no additional training of the model.
This approach is referred to as LLM-driven EA (LMEA).
Specifically, in each generation of the evolutionary search, LMEA instructs the LLM to select parent solutions from current population, and perform crossover and mutation to generate offspring solutions.
Then, LMEA evaluates these new solutions and include them into the population for the next generation.
LMEA is equipped with a self-adaptation mechanism that controls the temperature of the LLM.
This enables it to balance between exploration and exploitation and prevents the search from getting stuck in local optima.
We investigate the power of LMEA on the classical traveling salesman problems (TSPs) widely used in combinatorial optimization research.
Notably, the results show that LMEA performs competitively to traditional heuristics in finding high-quality solutions on TSP instances with up to 20 nodes.
Additionally, we also study the effectiveness of LLM-driven crossover/mutation and the self-adaptation mechanism in evolutionary search.
In summary, our results reveal the great potentials of LLMs as evolutionary optimizers for solving combinatorial problems.
We hope our research shall inspire future explorations on LLM-driven EAs for complex optimization challenges.
\end{abstract}

\begin{IEEEkeywords}
Evolutionary algorithms, large language model, combinatorial optimization, pre-trained model
\end{IEEEkeywords}

\section{Introduction}
Evolutionary algorithms (EAs) are a class of algorithms that draw inspiration from natural evolution~\cite{holland1992genetic}.
By simulating the principles of natural selection and genetic variation, EAs have been widely utilized in tackling complex combinatorial optimization problems arising in various domains,
including logistics~\cite{vidal2012hybrid,kocc2015hybrid,qi2015decomposition,FengOLT15}, cloud computing~\cite{zhan2015cloud,WangMCH22,yang2024reducing}, manufacturing~\cite{PanLW22,zhou2019self,zhang2022multitask}, robotics~\cite{starke2018memetic}, and adversarial example generation~\cite{liu2024effective}.
Despite the huge success achieved thus far, EAs, in general, still require carefully handcrafted operators to achieve high performance.
Typically, the design process of an EA involves algorithm experts analyzing the problem structure, tailoring specialized genetic operators (e.g., crossover and mutation) that exploit this structure most effectively, and continually refining these operators.
This process heavily relies on domain expertise and human efforts, and can become even more burdensome when dealing with new problems.

Motivated by the above challenge, recently there has been a surge of research interest in automating (part of) the design process of EAs, primarily through a meta-optimization paradigm~\cite{nwae132}. 
Specifically, these approaches construct a meta-level optimization problem where the decision variables are design choices of the algorithm, and the optimization objective is the algorithm's performance on a pre-collected set of problem instances (called the training set).
Then, by solving this meta-optimization problem, the design choices of the algorithm are automatically determined.
In the literature, there are various ways to represent the algorithm design choices, ranging from algorithm parameters~\cite{bezerra2015automatic,camacho2021pso}, search heuristics~\cite{feng2022towards,lin2017backtracking,burke2013hyper,drake2020recent}, algorithm selectors~\cite{ZhaoLYX2021}, to end-to-end deep neural networks~\cite{vinyals2015,bengio2021machine,liu2023good}.
However, the meta-optimization approaches still pose non-trivial challenges,
including how to select a representative training set that can sufficiently reflect the target cases where the algorithm will be applied~\cite{smith2015generating,TangLYY21},
how to build a compact and effective algorithm design space~\cite{BiedenkappLEHFH17},
and how to solve the meta-optimization problem~\cite{bello2016neural}.
Moreover, these approaches often need to consume significant computational resources to obtain a high-performing algorithm.

Large language models (LLMs) have recently yielded impressive results in a wide range of domains~\cite{min2023recent,lee2023benefits,thirunavukarasu2023large,kasneci2023chatgpt,liu2023summary}.
These models extract human knowledge by learning from vast amounts of text data and have demonstrated remarkable reasoning and decision-making capabilities~\cite{Wei0SBIXCLZ22,Wang2023,ZhouSHWS0SCBLC23,YaoZYDSN023}.
From this perspective, it is plausible that the knowledge embedded in LLMs also encompasses human experiences and intuitions in designing optimization algorithms.
This naturally raises an interesting question: can LLMs be used to help EAs solve complex optimization problems?

This work provides an affirmative answer to the above question.
Specifically, we propose an innovative approach, named LLM-driven EA (LMEA), for solving combinatorial optimization problems.
In each generation of the evolutionary search, LMEA constructs a prompt to instruct the LLM to select parent solutions from the current population, and perform crossover and mutation to generate offspring solutions.
Then, these new solutions are evaluated and added to the population for the next generation.
Additionally, a simple self-adaptation mechanism is integrated into LMEA to control the temperature of the LLM, thus balancing its exploration and exploitation.

From the perspective of designing EAs, LMEA has two appealing features.
First, due to the capabilities of LLMs, in LMEA we can describe the optimization problem and the desired solution properties in natural language to instruct the LLM.
In consequence, optimization with LMEA enables quick adaptation to different optimization problems by changing the problem description and solution specifications in the prompt.
Compared to the traditional practice of formally defining the problem and implementing operators through programming, LMEA follows an approach that is more direct and only demands minimal domain knowledge and human efforts.
Second, LMEA leverages the LLM in a zero-shot manner.
Here, the term ``zero-shot'' means that no additional training of the model is required, which is a significant advantage compared to meta-optimization approaches that require extensive computational resources to optimize the algorithm.

We investigate the power of LMEA on the classical traveling salesman problems (TSPs) widely used in combinatorial optimization research.
The experimental results demonstrate that, despite its minimal reliance on domain expertise, LMEA performs competitively to the traditional heuristics on TSP instances with up to 20 nodes.
Surprisingly, it consistently obtains the optimal solutions on TSP instances with 10 nodes and 15 nodes.
Furthermore, we conduct experiments to verify the effectiveness of the LLM-driven genetic operators and the self-adaptation mechanism.

To the best of our knowledge, this is the first attempt of utilizing LLMs in evolutionary combinatorial optimization.
Also, we would like to note that this work is not to show that LMEA can outperform those sophisticated specialized solvers for classical combinatorial optimization problems like TSPs.
Instead, the goal is to 
introduce an approach  with a significant departure from previous design paradigms of EAs.
Importantly, this approach indeed demonstrates its capacity to solve non-trivial hard combinatorial optimization problems.
We hope that our results will inspire further exploration of LLM-driven EAs for combinatorial optimization challenges.

The remainder of this paper is organized as follows.
Section~\ref{sec:related_work} briefly reviews the literature on EAs for combinatorial optimization, LLMs and prompts, as well as the intersection between EAs and LLMs.
Section~\ref{sec:method} presents the approach LMEA.
Experiments on TSPs are presented in Section~\ref{sec:exp}.
Finally, Section~\ref{sec:conclu} concludes the paper with discussions.

\section{Related Works}
\label{sec:related_work}

\subsection{EAs for Combinatorial Optimization}
Combinatorial optimization involves finding the best possible solution from a finite solution set.
It has many applications in various domains~\cite{korte2011combinatorial}.
From the perspective of computational complexity, many combinatorial optimization problems are NP-hard due to their discrete and nonconvex nature~\cite{hochba1997approximation}.
For these problems, exact methods such as the branch and bound algorithms~\cite{lawler1966branch} can find the optimal solutions, but generally suffer from exponential time complexity.
In contrast, meta-heuristics seek to find good (but not necessarily optimal) solutions within reasonable computation time.
EAs, as one of the mainstream meta-heuristics, have achieved significant progress in solving combinatorial optimization problems over the past decades and are still undergoing diverse and flourishing development.
Currently, EAs have established themselves as state-of-the-art methods for various combinatorial optimization problems \cite{vidal2012hybrid,kocc2015hybrid,qi2015decomposition,FengOLT15,zhan2015cloud,WangMCH22,PanLW22,zhou2019self,zhang2022multitask,starke2018memetic}. 
These algorithms range from classic EAs like genetic algorithms~\cite{vidal2012hybrid,kocc2015hybrid}, hybrid approaches such as memetic algorithms~\cite{FengOLT15,qi2015decomposition,WangMCH22,starke2018memetic}, to co-evolutionary algorithms~\cite{chaabani2018new}.

In general, when applying EAs to solve combinatorial optimization problems, algorithm practitioners often need to design operators tailored to the problem to obtain satisfactory performance.
These operators can be specific to the solution representations, such as the various crossover variants~\cite{umbarkar2015crossover} (e.g., single-point, multi-point, shuffle, and matrix) for binary solution representations in set maximization problems, as well as the k-opt mutation~\cite{helsgaun2009general} and edge assembly crossover~\cite{nagata2013powerful} for the permutation-based solution representations in TSPs and vehicle routing problems (VRPs).
In certain cases, one also needs to design repair operators~\cite{salcedo2009survey} to ensure that the solutions found by EAs adhere to problem constraints, and regrouping operators~\cite{vidal2012hybrid} to maintain population diversity.
Overall, the design of effective operators relies on a deep understanding of the problem of interest, which requires extensive domain expertise and human efforts~\cite{liu2023good}.

\subsection{Large Language Models (LLMs) and Prompts}
LLMs are large deep neural networks (often with billions of parameters) which are trained on vast amounts of text data to perform next-token prediction, i.e., predict the proceeding token given a sequence of tokens seen before.
In the last two years, scaling up LLMs has yielded groundbreaking performance across a broad spectrum of tasks~\cite{min2023recent,zhao2023survey,tian2023chatgpt,lee2023benefits,blocklove2023chip,zheng2023can,thirunavukarasu2023large,kasneci2023chatgpt,liu2023summary}.
Among them of particular interest are tasks involving reasoning and decision-making, such as planning~\cite{HuangAPM22,xi2023rise,YaoZYDSN023} and solving mathematical problems~\cite{LewkowyczADDMRS22,ZhouSHWS0SCBLC23,Wang2023}.
Given the inherent inter-connections between optimization, reasoning, and decision-making, it is likely that LLMs could also demonstrate competence in optimization tasks, which is indeed one motivation behind this work.

A prompt is the instruction that guides LLMs into generating desired output.
Although a prompt can be in various forms such as a question, a sentence, or a keyword, there has been much evidence~\cite{Wei0SBIXCLZ22,Wang2023,ZhouSHWS0SCBLC23} showing that the prompt format can significantly influence the quality of the LLM's output.
In general, when faced with a specific task, one needs to carefully craft the prompt to align with the desired outcome.
This involves considering the context, the level of detail required, and the clarity of the instruction.
Previous research has identified several effective techniques for prompting.
For example, one can include several task examples as demonstrations in the prompt, i.e., in-context learning~\cite{BrownMRSKDNSSAA20}, and can explicitly instruct the LLM to think step by step, i.e.,  chain of thoughts ~\cite{Wei0SBIXCLZ22}.
In this work, to make LLMs function effectively as the operators of EAs, we carefully design a well-structured prompt that consists of several parts including task descriptions, solution properties, 
population information, and operator instructions. 

\subsection{Intersection between EAs and LLMs}
Currently, research at the intersection of LLMs and EAs is still in its early stage.
Due to the remarkable capabilities of LLMs in code generation, several recent works have attempted to combine LLMs with EAs to generate neural network structures~\cite{chen230214838}, programs that fulfill specific functionalities~\cite{Lehan2206}, and meta-heuristics~\cite{PluhacekKKVS23}.
Furthermore, the core ideas of EAs, including natural selection and genetic variations, have also been employed to evolve prompts to enhance the performance of LLMs~\cite{Guo2309}.
Regarding directly applying LLMs to solve optimization problems, there is very limited research.
One such work is~\cite{yang2023large}, which presents preliminary results of using LLMs for solving linear regression problems and TSPs.
Another very recent work~\cite{liu2023large} incorporates LLMs as the mutation operator in continuous multi-objective optimization algorithm.
However, the utilization of LLMs for evolutionary combinatorial optimization still remains unexplored.

\begin{figure*}[tbp]
	\centering
	\scalebox{0.98}{
		\includegraphics[width=\textwidth]{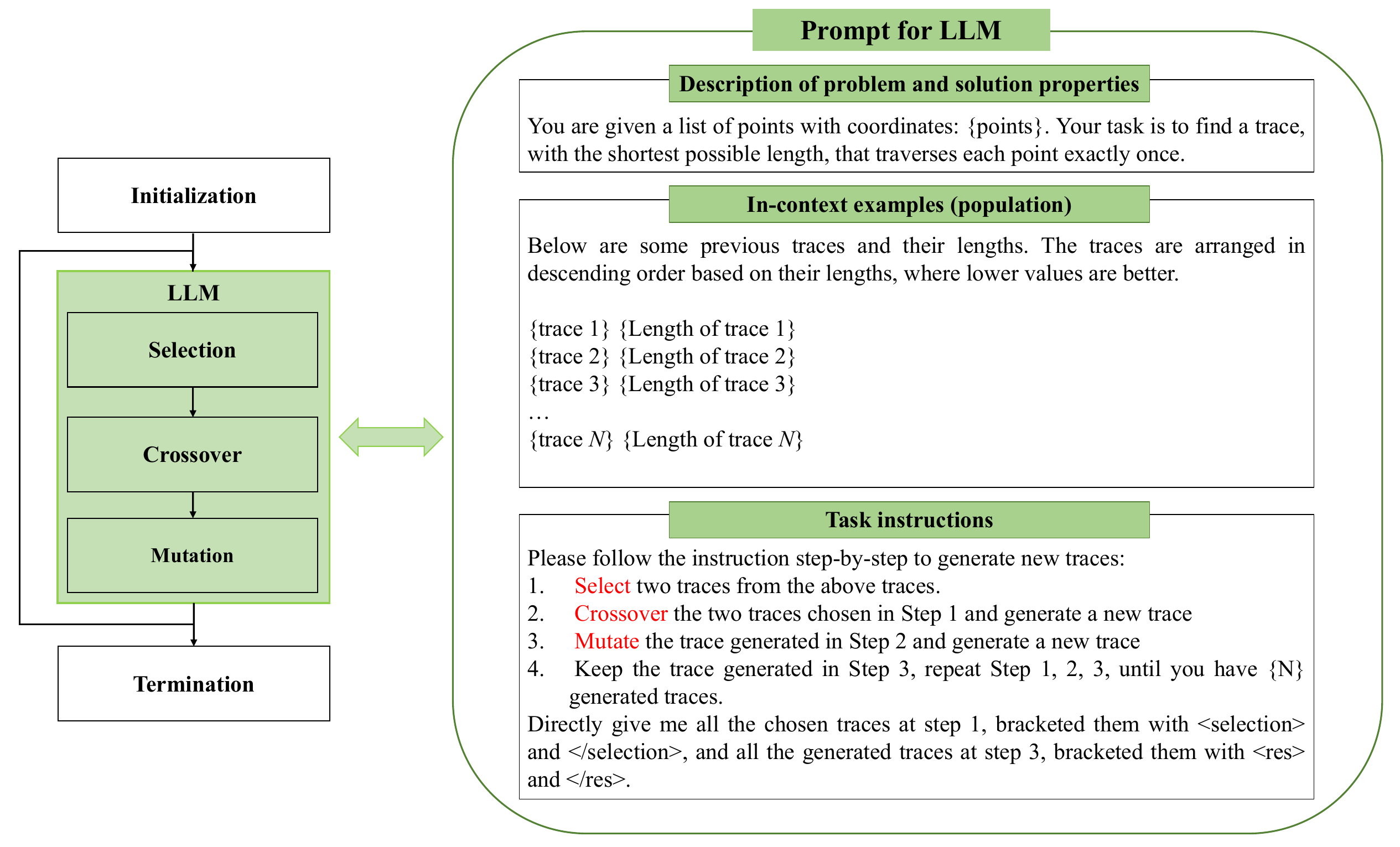}}
	\caption{An overview of LMEA. The right half of this diagram demonstrates an example of the constructed prompt when utilizing LMEA to solve TSPs.
		The contents within ``\{\}'' in the prompt will be replaced with the corresponding input.}
	\label{fig:LMEA}
\end{figure*}

\begin{algorithm}[tbp]
\LinesNumbered
\KwIn{
The optimization problem $T$,
maximum number of generations $G$,
population size $N$;
}
\KwOut{the best found solution $s^*$}
$P \leftarrow$ randomly initialize $N$ solutions to $T$;\\
$g=1$;\\
\While{$g \leq G$}
{
$prompt \leftarrow$ construct prompt based on  $T$ and $pop$;\\
$P' \leftarrow $ instruct LLM with $prompt$ to generate $N$ offspring solutions;\\
$P \leftarrow $ the top $N$ solutions among $P \cup P'$;\\
Self-adapt the temperature of LLM if necessary;\\
$g \leftarrow g+1$;\\
}
$s^* \leftarrow$ the best solution in $P$;\\
\Return{$s^*$}
\caption{LLM-driven EA (LMEA)}
\label{alg:LMEA}
\end{algorithm}

\section{LMEA for Combinatorial Optimization}
\label{sec:method}
In this section, we first present the framework of LMEA, then delve into the construction of prompts to instruct the LLM, and finally describe the self-adaptation mechanism of the LLM's temperature.
An overview of LMEA is also illustrated in Figure~\ref{fig:LMEA}.

\subsection{Algorithm Framework}
As presented in Algorithm~\ref{alg:LMEA}, LMEA follows the conventional EA framework~\cite{holland1992genetic}.
Given an optimization problem $T$, LMEA first randomly generates $N$ solutions to form the initial population (line 1) and then proceeds with the evolutionary process (lines 3-9).
In each generation, LMEA constructs a prompt (line 4) to guide LLM in selecting parent solutions from the current population and then conduct crossover and mutation based on them, generating $N$ offspring solutions (line 5).
These $N$ solutions are then combined with the current population (line 6) and the top $N$ solutions are retained for the next generation (line 7). 
In addition, the LLM's temperature would be adjusted if necessary (see Section~\ref{sec:self-adapt}).
Finally, LMEA would terminate when the number of generations reaches a predefined number and the best found solution is returned (lines 10-11).

\subsection{LLMs as Evolutionary Optimizers}
LMEA employs the LLM as evolutionary operators in a zero-shot manner.
Specifically, the parent selection and genetic variations (crossover and mutation) are accomplished through the in-context learning process of the LLM~\cite{BrownMRSKDNSSAA20}, which is facilitated by carefully constructed prompts.

To be concrete, the prompt consists of three parts:
\begin{itemize}
	\item \textbf{Problem description and solution properties}: this part includes the description of the optimization problem to be solved and specifications of the desired solution properties.
	\item \textbf{In-context examples}: some solutions to the optimization problem and their corresponding fitness are provided as demonstrations for the LLM. Typically, these solutions are derived from the current population.
	\item \textbf{Task Instructions}: this part provides explicit instructions for the LLM to perform parent selection and carry out crossover and mutation, generating new solutions.
\end{itemize}
Figure~\ref{fig:LMEA} illustrates an example of the constructed prompt when using LMEA to solve TSPs.
The problem description contains the coordinates of the points in the TSP instance, while the solution properties specify the constraints that TSP solutions must meet (traversing each point exactly once) and that shorter lengths are preferable.
In-context examples consist of TSP solutions from the current population and their corresponding lengths.
Task Instructions guide the LLM in generating new TSP solutions.

It is important to note that, unlike the traditional practice of implementing evolutionary operators step-by-step through programming, LMEA does not instruct the LLM on how to precisely perform parent selection, crossover, and mutation.
Instead, LMEA instructs the LLM at a higher level using natural language.
This approach only requires minimal reliance on domain expertise.
Finally, the prompt also strictly defines the format of the LLM's output to enable LMEA to interpret the output easily.
For example, the results of parent selection are enclosed between $<$selection$>$ and $<$/selection$>$, and the generated solutions are enclosed between $<$res$>$ and $<$/res$>$.

\subsection{Self-Adaptation of the LLM's Temperature}
\label{sec:self-adapt}
The temperature of LLMs, which is used in the sampling process when generating text, is a parameter that controls the randomness or entropy of the text~\cite{BrownMRSKDNSSAA20}. 
A higher temperature value increases the randomness, while a lower value makes the model's output more deterministic.

From the perspective of search-based optimization, the temperature of LLMs can be understood as a parameter that controls the exploration in the search process.
A higher temperature equips the LLM with stronger exploratory ability.
Based on this, we propose a simple rule for adaptively adjusting the temperature: If LMEA fails to find a solution better than the current best for $K$ consecutive generations, the temperature value will be increased by $\alpha$.
Typically, the default temperature of LLMs is 1.0~\cite{BrownMRSKDNSSAA20}; in this work, we always set $K=20$ and $\alpha=0.1$.

\section{Experiments}
\label{sec:exp}
We investigate the power of LMEA on TSPs.
Specifically, the experiments aim to address the two questions below.\footnote{Code and dataset is available at \url{https://github.com/cschen1205/LMEA}}
\begin{itemize}
\item How good is LMEA's performance, especially compared to those traditional hand-designed heuristics?
\item Are LLM-driven genetic operators and the self-adaptation mechanism useful in improving the optimization performance?
\end{itemize}

\subsection{Test Problem Instances}
We considered EUC-2D TSPs, where the nodes are defined on a two-dimensional plane and the distances between two nodes are the same in both directions.
Specifically, two different types of TSP instances were generated through the generators which has been used to create testbeds for the 8-th DIMACS Implementation Challenge~\cite{gutin2006traveling}.\footnote{Generators available at: \url{http://dimacs.rutgers.edu/archive/Challenges/TSP}}
\begin{itemize}
	\item The \textit{portgen} generator generates a TSP instance (called a rue instance) by uniformly and randomly placing nodes on a two-dimensional plane.
	\item The  \textit{portcgen}  generator generates a TSP instance (called a clu instance) by randomly placing nodes around different central nodes.
\end{itemize}
For both types (rue and clu), the instances were generated such that both $x$ and $y$ coordinates of the nodes lie within [0, 100].
For each instance type, we considered four different problem sizes (number of nodes, denoted as $n$), i.e., $n=10, 15, 20, 25$.
In summary, there were eight different combinations of instance types and problem sizes.
We denote them as rue/clu-10/15/20/25, respectively, and for each of them, we generated five TSP instances.
For all the 40 generated instances, Concorde~\cite{applegate2006concorde}, an exact TSP solver, was used to obtain their optimal solutions.\footnote{Concorde available at \url{https://www.math.uwaterloo.ca/tsp/concorde.html}} 

\begin{table*}[tbp]
	\centering
	\caption{Test results on eight test sets with different numbers of nodes ($n=10, 15, 20, 25$) and TSP types (rue and clu). Each test set contains 5 TSP instances, and the average optimality gap (\%) $\pm$ standard deviation achieved on them is reported.  On each test set, the best average optimality gap is indicated in \textbf{bold}.
		``\# Generations'' represents the mean $\pm$ standard deviation of the generation numbers that LMEA and OPRO finds the optimal solution. ``\# Successes'' counts the number of TSP instances that  LMEA and OPRO finds the optimal solution. ``N/A'' means that no optimal solution is found for any TSP instance in the test set.}
	\begin{tabular}{ccccccccc}
		\toprule
		\multirow{2}[4]{*}{Test set} & \multicolumn{6}{c}{Optimality gap (\%)}       & \multicolumn{2}{c}{\# Generations (\# Successes)} \\
		\cmidrule(rl){2-7}   \cmidrule(rl){8-9}       & NN    & FI    & NI    & RI    & LMEA  & OPRO  & LMEA  & OPRO\\
		\midrule
		rue-10 & 11.22 $\pm$ 3.35 & 2.23 $\pm$ 1.26 & 0.58 $\pm$ 0.58 & \textbf{0.00 $\pm$ 0.00} & \textbf{0.00 $\pm$ 0.00} & \textbf{0.00 $\pm$ 0.00} & 35.80 $\pm$ 7.17 (5) & 60.60 $\pm$ 13.68 (5) \\
		rue-15 & 9.84 $\pm$ 3.34 & 1.08 $\pm$ 1.01 & 0.79 $\pm$ 0.79 & 2.45 $\pm$ 1.18 & \textbf{0.06 $\pm$ 0.06} & 5.23 $\pm$ 2.01 & 235.25 $\pm$ 6.12 (4) & 189.00 $\pm$ 0.00 (1) \\
		rue-20 & 21.47 $\pm$ 2.01 & \textbf{1.99 $\pm$ 0.86} & 2.65 $\pm$ 1.43 & 2.15 $\pm$ 1.26 & 3.94 $\pm$ 1.54 & 26.30 $\pm$ 3.58 & 197.00 $\pm$ 0.00 (1) & N/A (0) \\
		rue-25 & 10.71 $\pm$ 3.36 & 2.33 $\pm$ 1.34 & \textbf{1.41 $\pm$ 0.72} & 2.41 $\pm$ 0.74 & 18.72 $\pm$ 3.31 & 53.59 $\pm$ 8.37 & N/A (0) & N/A (0) \\
		clu-10 & 16.48 $\pm$ 2.02 & 1.28 $\pm$ 0.79 & 0.99 $\pm$ 0.99 & 1.37 $\pm$ 0.84 & \textbf{0.00 $\pm$ 0.00} & \textbf{0.00 $\pm$ 0.00} & 19.00 $\pm$ 3.30 (5) & 90.40 $\pm$ 16.55 (5) \\
		clu-15 & 23.39 $\pm$ 4.39 & \textbf{0.00 $\pm$ 0.00} & 0.48 $\pm$ 0.48 & 0.08 $\pm$ 0.08 & 0.11 $\pm$ 0.11 & 8.13 $\pm$ 4.83 & 152.25 $\pm$ 18.80 (4) & 153.00 $\pm$ 0.00 (1) \\
		clu-20 & 21.29 $\pm$ 2.77 & \textbf{0.78 $\pm$ 0.48} & 3.81 $\pm$ 1.45 & 1.97 $\pm$ 0.87 & 4.05 $\pm$ 0.69 & 19.83 $\pm$ 4.76 & N/A (0) & N/A (0) \\
		clu-25 & 22.36 $\pm$ 1.29 & 2.10 $\pm$ 0.55 & 3.58 $\pm$ 0.68 & \textbf{1.83 $\pm$ 0.62} & 10.06 $\pm$ 1.69 & 48.25 $\pm$ 5.86 & N/A (0) & N/A (0) \\
		\bottomrule
	\end{tabular}
	\label{tab:main_results}
\end{table*}

\subsection{Baseline Algorithms}
We considered the following four traditional heuristics for solving TSPs as baseline algorithms.\footnote{Implementations available at \url{https://github.com/Valdecy/pyCombinatorial}}
\begin{itemize}
	\item \textbf{Nearest neighbor (NN).} This heuristic first randomly chooses a node which is the starting point of the tour.
	Then at each step, the next node is chosen as the one nearest to the last node of the tour and is appended to the tour as the new last node.
	This process finishes when the tour contains all the nodes.
	
	\item \textbf{Farthest/nearest/random insertion (FI/NI/RI).} The insertion heuristics minimize the cost of inserting a node into the tour.
	At each step, for each node $k$, its position of insertion is determined such that the cost $c(k)=d_{i,k}+d_{k,j}-d_{i,j}$ is minimized, where $i$ and $j$ are adjacent nodes in the current tour and $d$ indicates the distance.
	The different variants of insertion heuristics differ in how the inserted node is selected.
	FI selects the node with the largest distance to any node of the tour.
	NI selects the node that is nearest to any node in the tour.
	RI selects a random node.
\end{itemize}

Additionally, we considered the very recent approach Optimization by PROmpting (OPRO)~\cite{yang2023large} as the baseline.
OPRO is also driven by LLMs, but it differs from LMEA in that OPRO does not employ LLM to perform crossover and mutation.
Instead, in each generation, OPRO directly instructs the LLM to generate $N$ new solutions based on the current population.
Therefore, OPRO can be viewed as a variant of LMEA without LLM-driven genetic operators (crossover and mutation).
The comparison between them can validate the effectiveness of LLM-driven genetic operators.

\subsection{Experimental Setup}
To make a fair comparison, for LMEA and OPRO, the population size $N$ and maximum generation number $G$ were set the same, i.e.,  $N=16$ and $G=250$.
For LMEA, we used the \emph{chat-turbo-0613} version of the GPT-3.5 API as the LLM.\footnote{Details of the API available at: \url{https://platform.openai.com/}}
On each test set, i.e., rue/clu-10/15/20/25, we executed each algorithm and reported its average optimality gap on the set.
The optimality gap is defined as the difference between the length of the best solution found by the algorithm (denoted as $len(s^*)$) and the length of the optimal solution (denoted as $opt$), calculated using $(len(s^*) - opt) / opt$.

\begin{table}[tbp]
	\centering
	\caption{Comparison of LMEA and its variant without self-adaptation (LMEA*) on the rue-20 test set.}
	\begin{tabular}{ccc}
		\toprule
		\multirow{2}[4]{*}{Test set} & \multicolumn{2}{c}{Optimality gap (\%)} \\
		\cmidrule{2-3}          & LMEA  & LMEA* \\
		\midrule
		rue-20 & 3.94 $\pm$ 1.54 & 11.82 $\pm$ 2.21 \\
		\bottomrule
	\end{tabular}
	\label{tab:self_adaptation}
\end{table}

\subsection{Results and Analysis}
The test results are presented in Table~\ref{tab:main_results}.
In addition to the optimality gap, the numbers of generations needed by LMEA and OPRO to find the optimal solutions are also reported.

First, it can be observed that on TSP instances with 10 nodes and 15 nodes, LMEA outperforms the heuristics on three out of four test sets, i.e., rue-10/15 and clu-10.
Taking a closer look, on the rue/clu-10/15 test sets, it is interesting to find that LMEA consistently finds the optimal solution on 19 out of 20 instances.
Considering the size of the solution space of a rue/clu-15 TSP instance is at least in the order of billions, LMEA can find the optimal solution to it within a total fitness evaluation number not exceeding $250 \times 16=4000$ (note that $G=250$ and $N=16$).
This clearly shows that, despite its minimal reliance on domain expertise, LMEA has the capability to optimize non-trivial NP-hard combinatorial optimization problems like TSPs.

Second, on TSP instances with 20 nodes, LMEA performs slightly worse than the heuristics (except NN).
As the node number increases further ($n=25$), the optimality gap of LMEA increases rapidly.
Addressing the scalability limitations of LMEA is a crucial area for future research.

\begin{figure*}[tbp]
	\centering
	\subfloat[rue-10]
	{
			\includegraphics[width=.49\columnwidth]{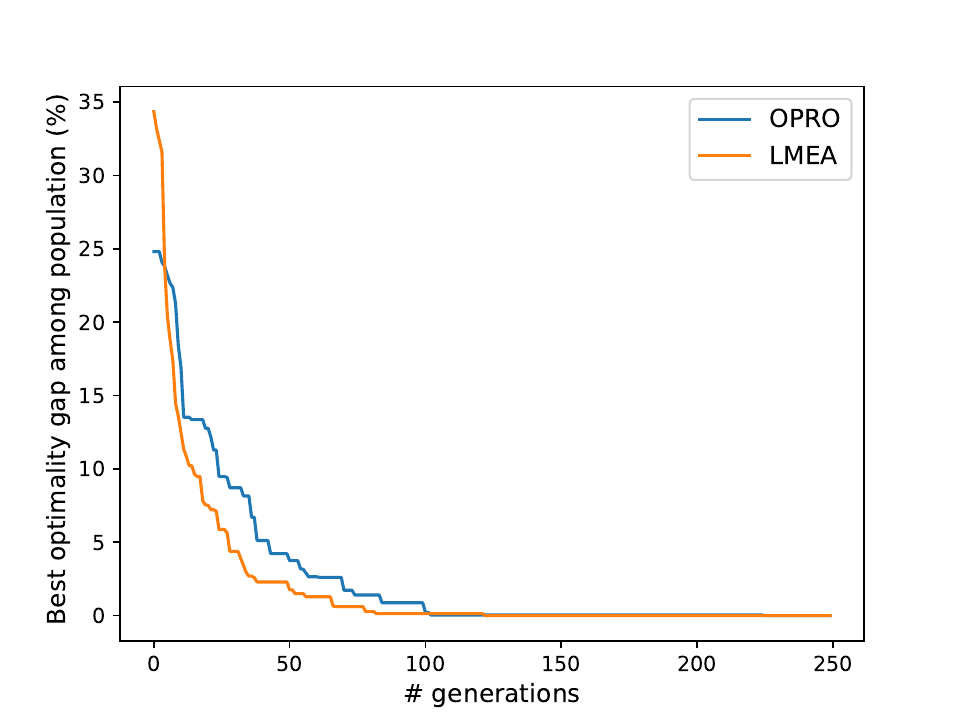}
			\includegraphics[width=.49\columnwidth]{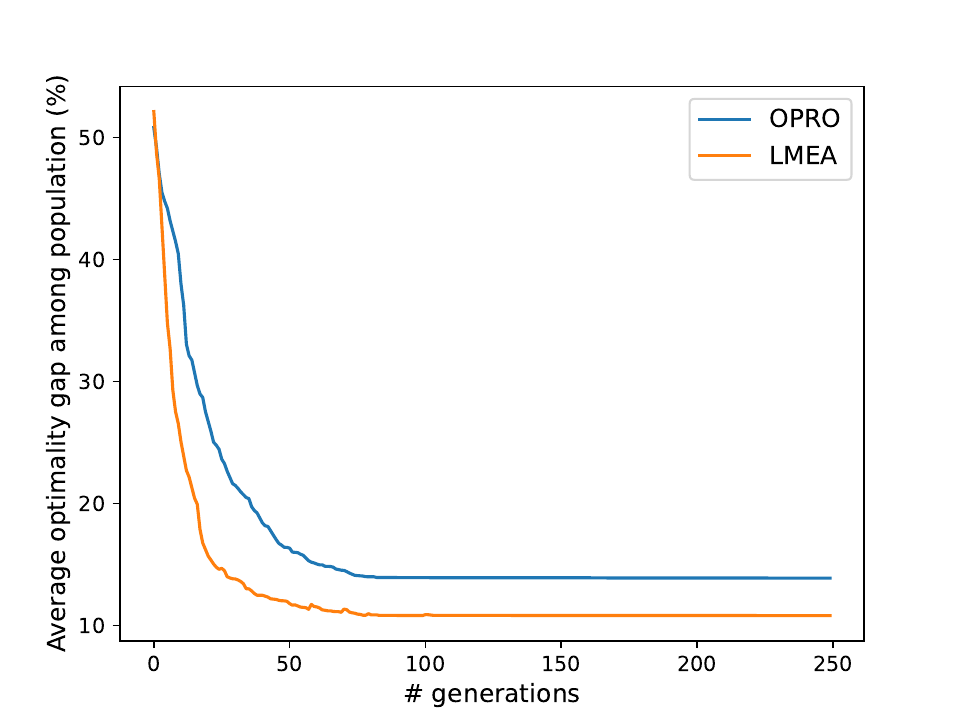}
		}
	\subfloat[clu-10]{
			\includegraphics[width=.49\columnwidth]{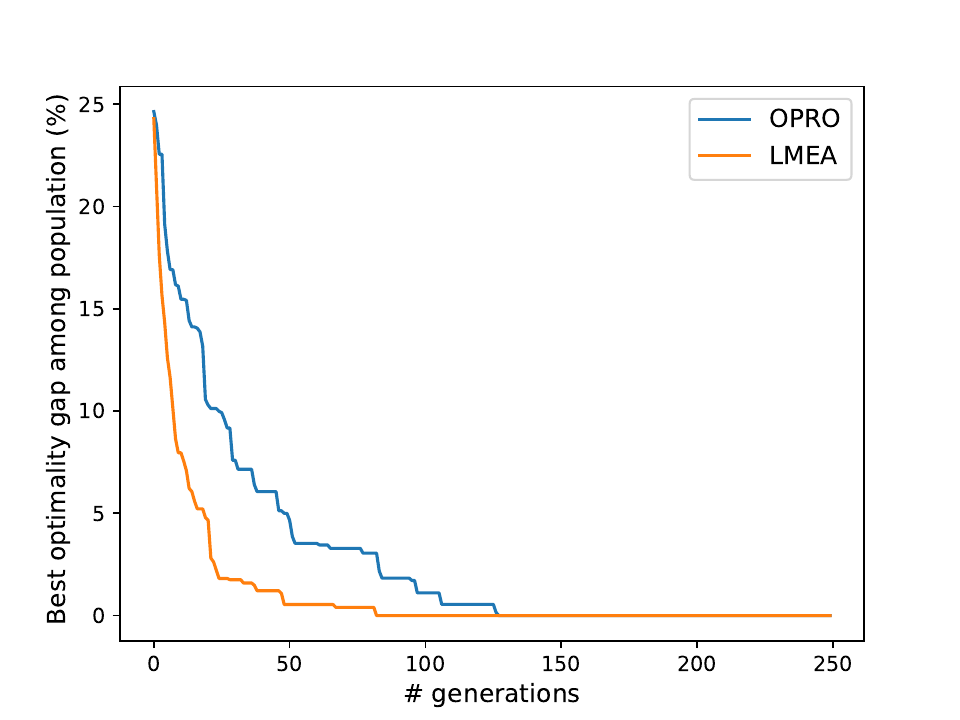}
			\includegraphics[width=.49\columnwidth]{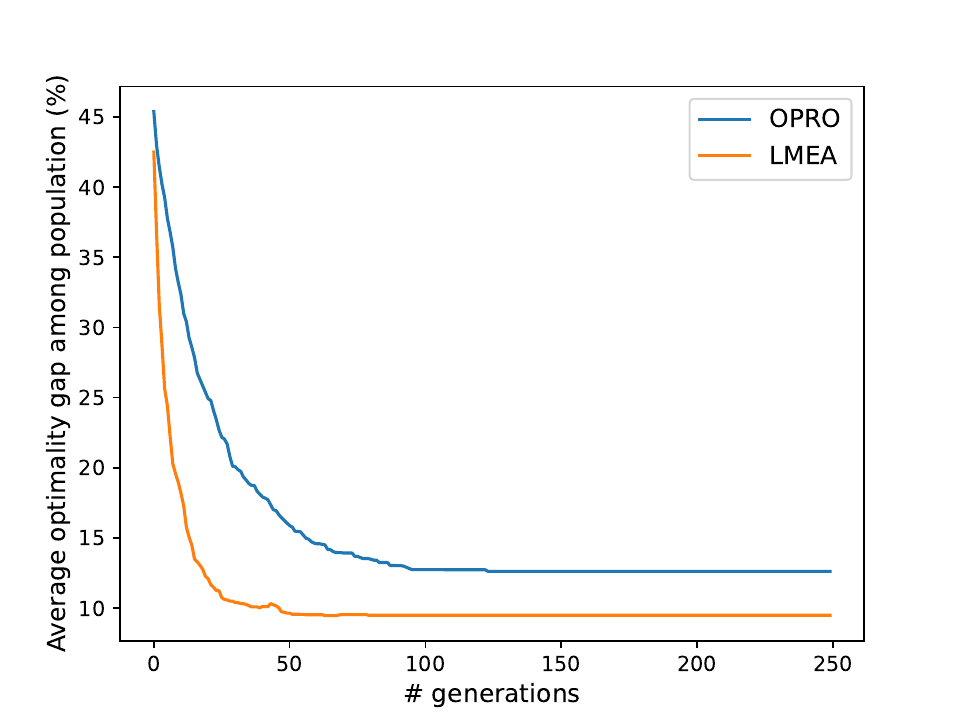}
		}
	\hfil
	
	\subfloat[rue-15]
	{
			\includegraphics[width=.49\columnwidth]{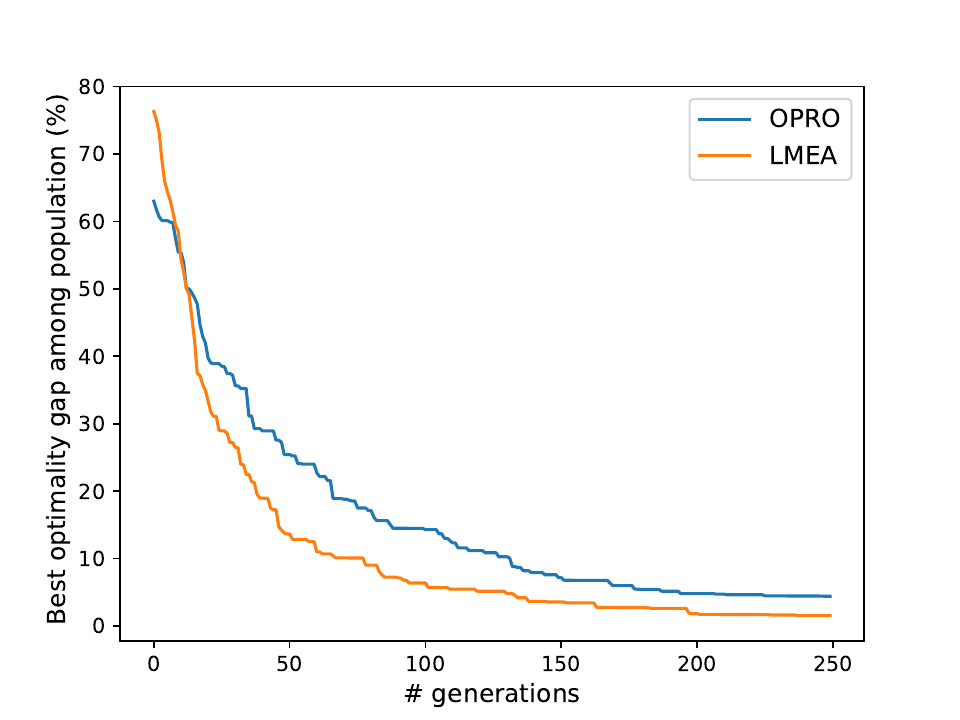}
			\includegraphics[width=.49\columnwidth]{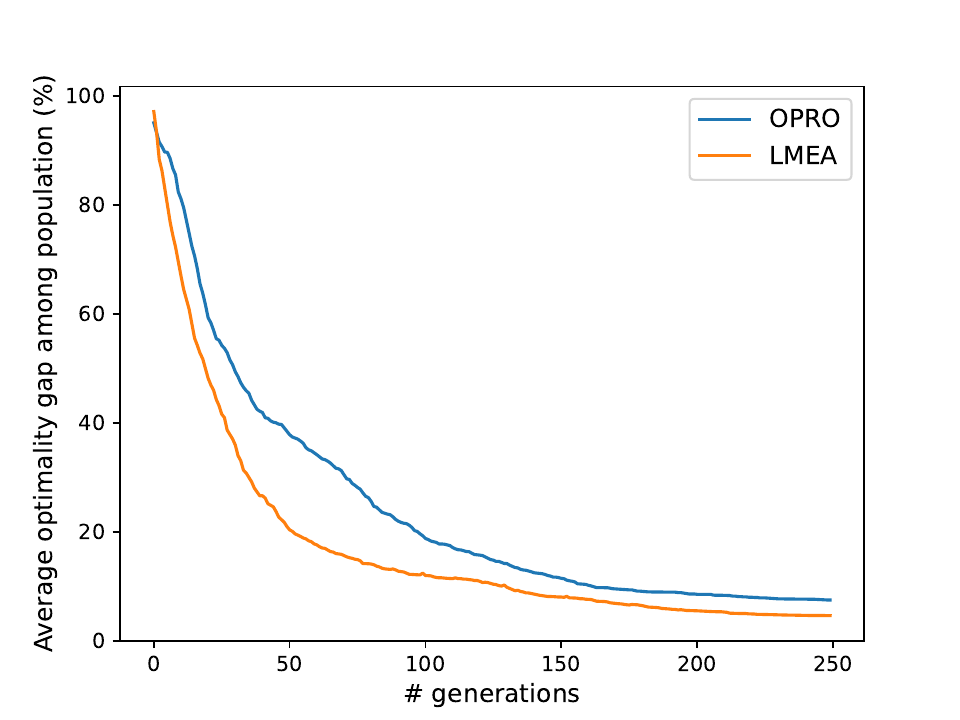}
		}
	\subfloat[clu-15]{
			\includegraphics[width=.49\columnwidth]{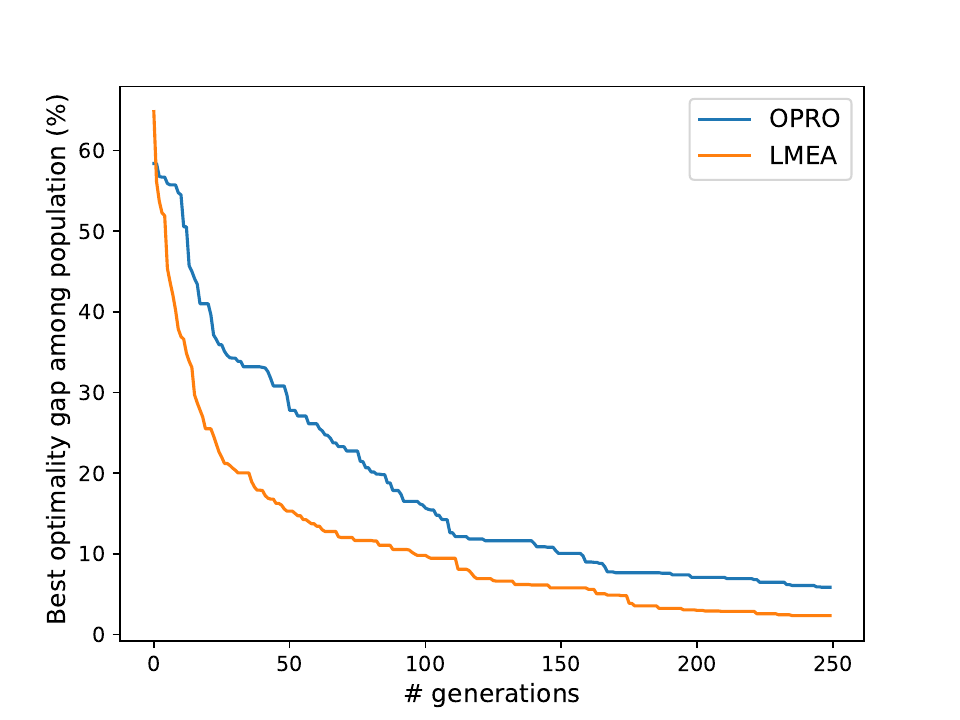}
			\includegraphics[width=.49\columnwidth]{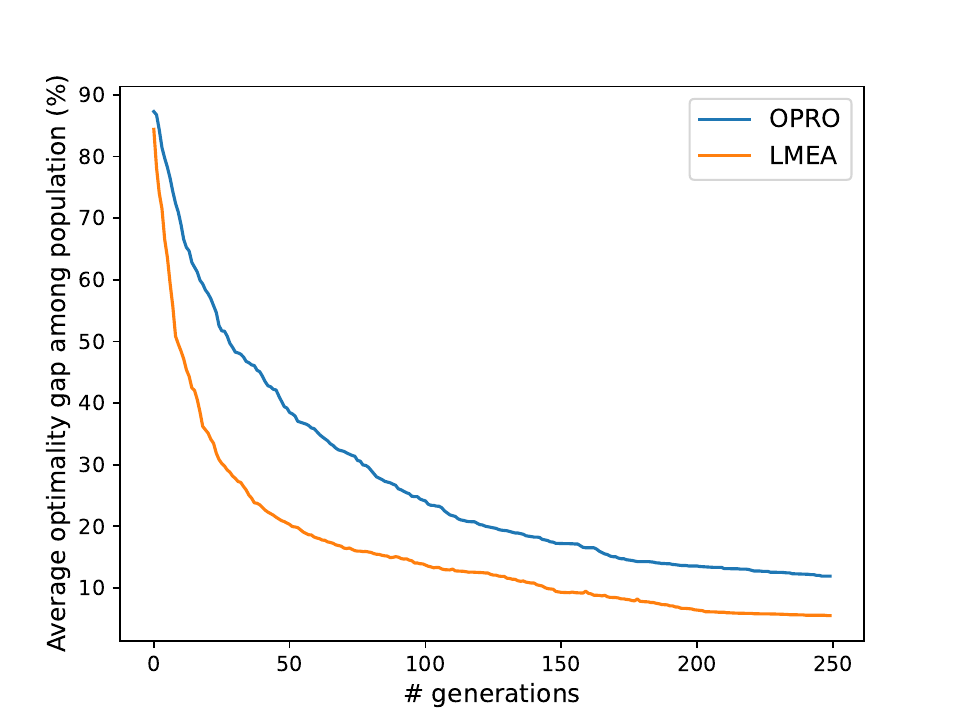}
		}
	\hfil
	
	\subfloat[rue-20]
	{
			\includegraphics[width=.49\columnwidth]{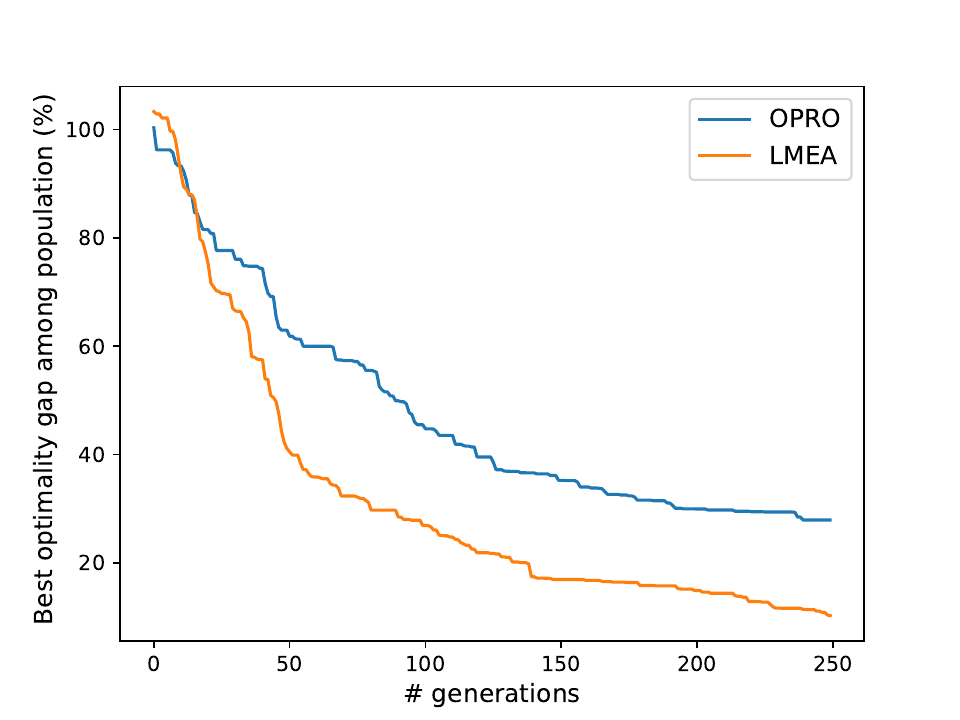}
			\includegraphics[width=.49\columnwidth]{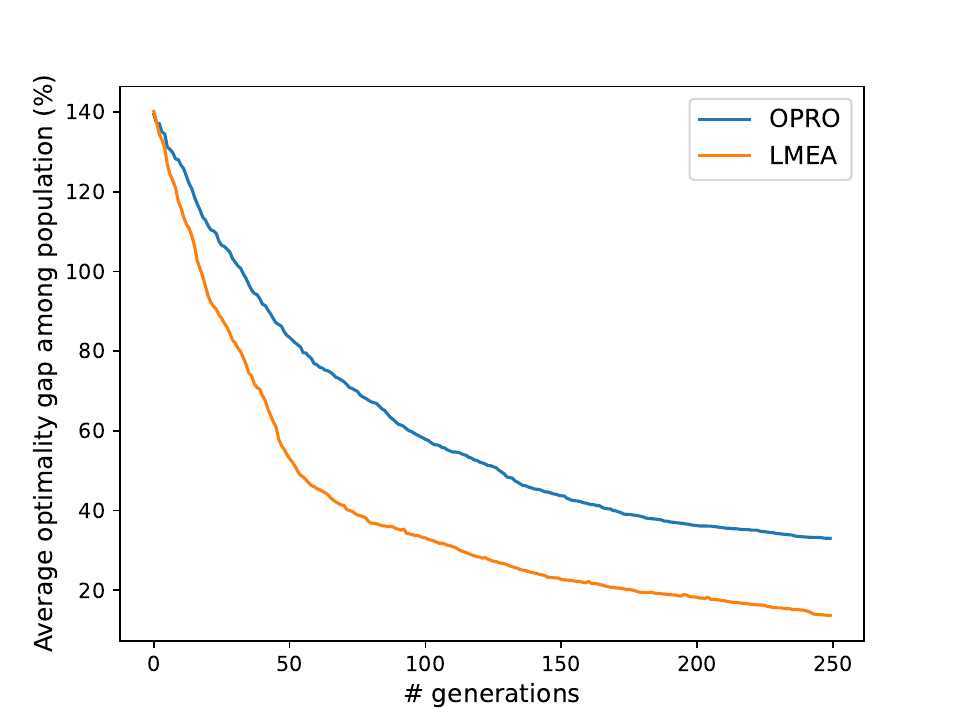}
		}
	\subfloat[clu-20]{
			\includegraphics[width=.49\columnwidth]{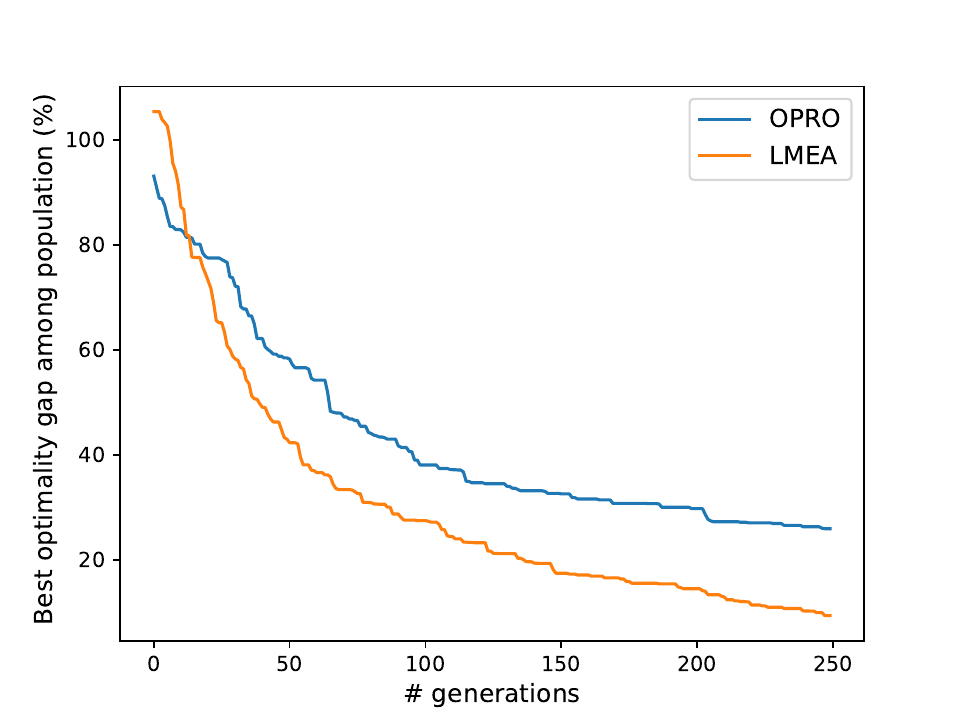}
			\includegraphics[width=.49\columnwidth]{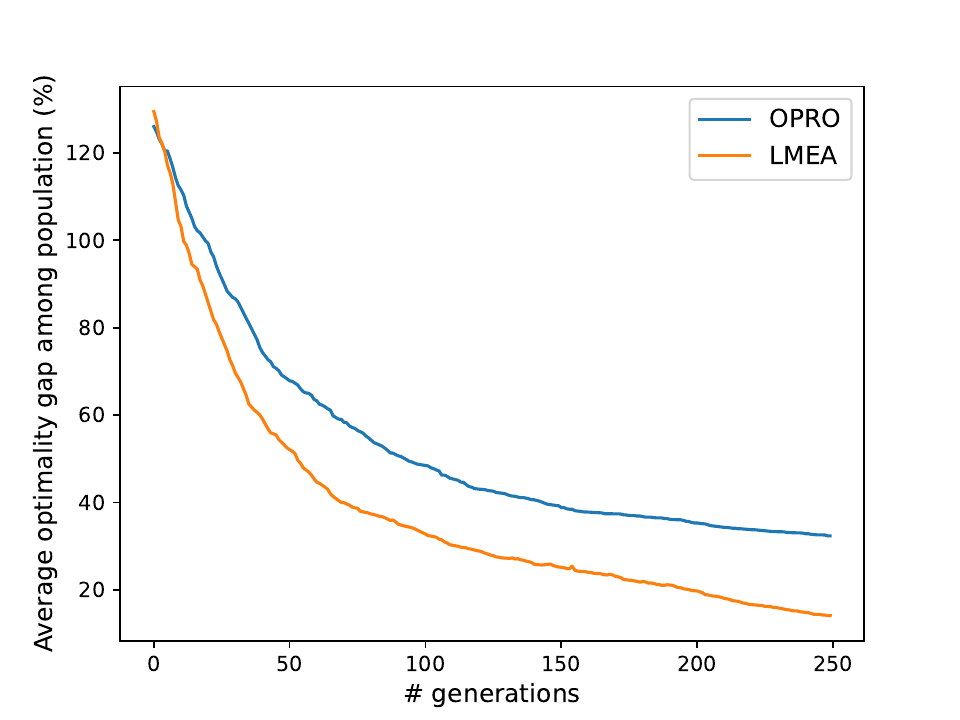}
		}
	\hfil
	
	\subfloat[rue-25]
	{
			\includegraphics[width=.49\columnwidth]{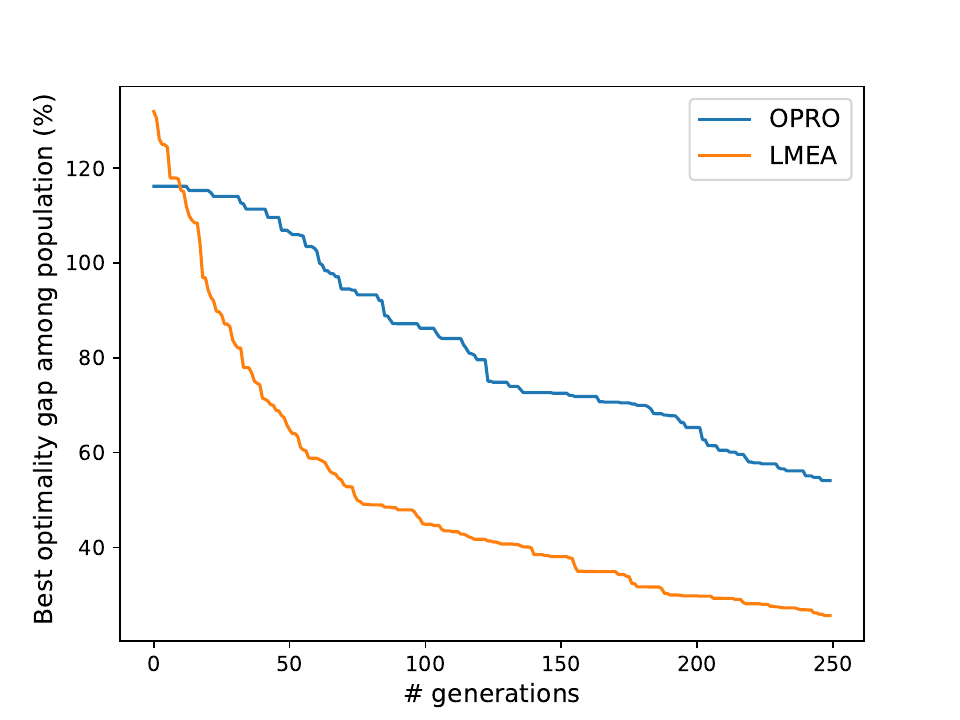}
			\includegraphics[width=.49\columnwidth]{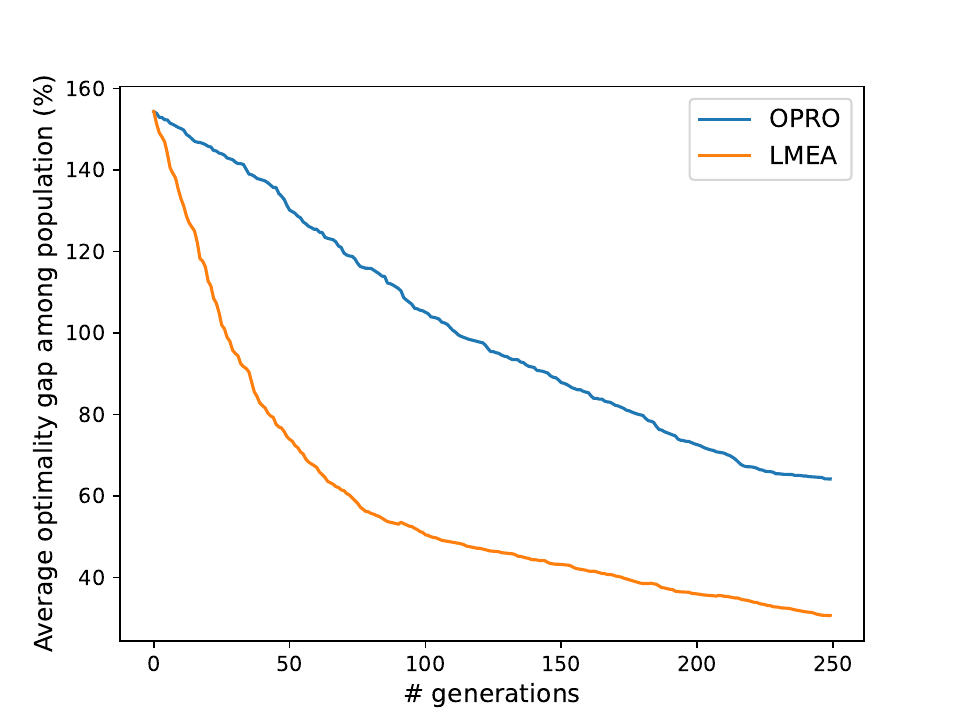}
		}
	\subfloat[clu-25]{
			\includegraphics[width=.49\columnwidth]{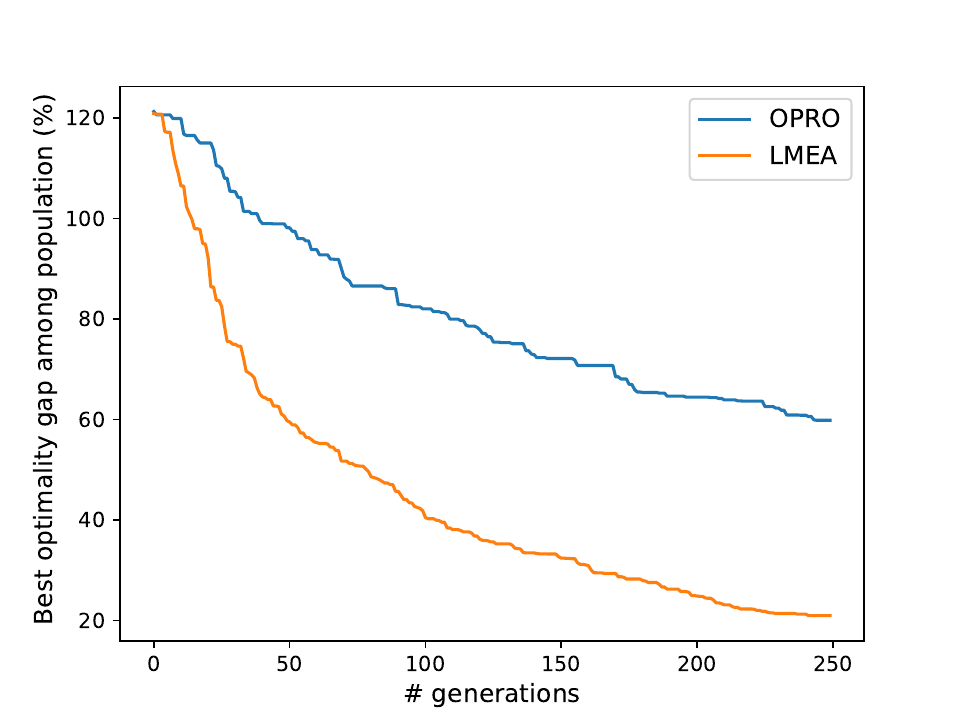}
			\includegraphics[width=.49\columnwidth]{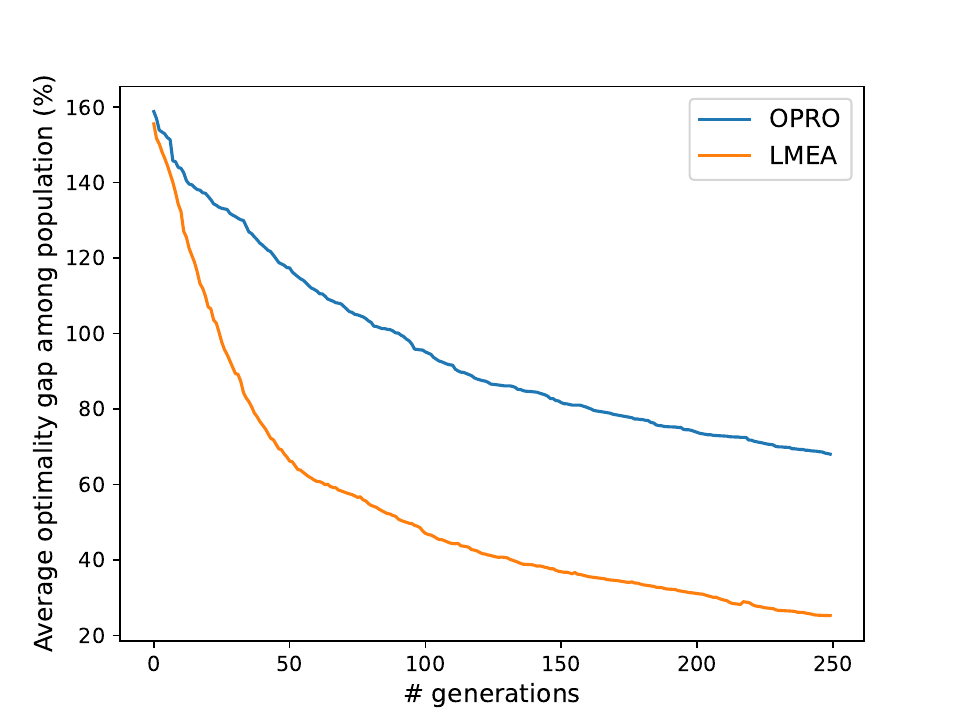}
		}
	
	\caption{Convergence curves: optimality gaps achieved by LMEA and OPRO as generation number increases.
		For each test set, the \textbf{left} figure illustrates the average optimality of the population, and the \textbf{right} figure illustrates optimality gap of the best found solution
	}
	\label{fig:convergen_curves}
\end{figure*}

Finally, on all the eight test sets, LMEA outperforms OPRO, and the performance gap becomes larger as the node number increases.
Figure~\ref{fig:convergen_curves} also illustrates the convergence curves of LMEA and OPRO on all the test sets.
It can be observed that in general LMEA can find better solutions more quickly than OPRO.
Since OPRO can be viewed as a variant of LMEA without LLM-driven crossover and mutation, these results have confirmed the effectiveness of the LLM-driven genetic operators.



\begin{figure}[tbp]
\centering
\subfloat[]{\includegraphics[width=.5\columnwidth]{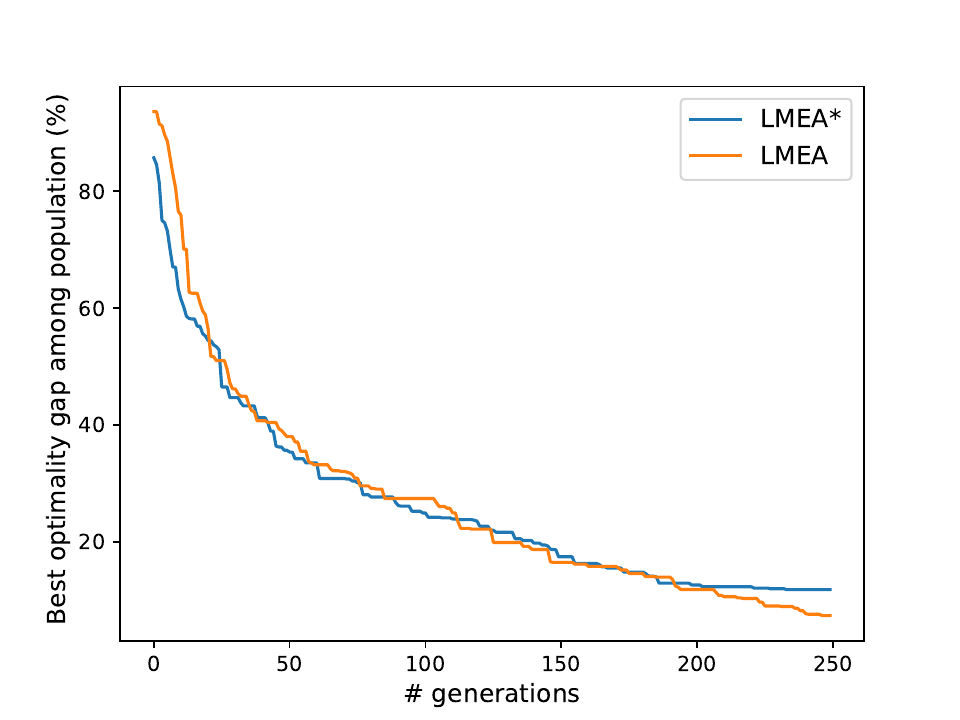}}
\hfil
\subfloat[]{\includegraphics[width=.5\columnwidth]{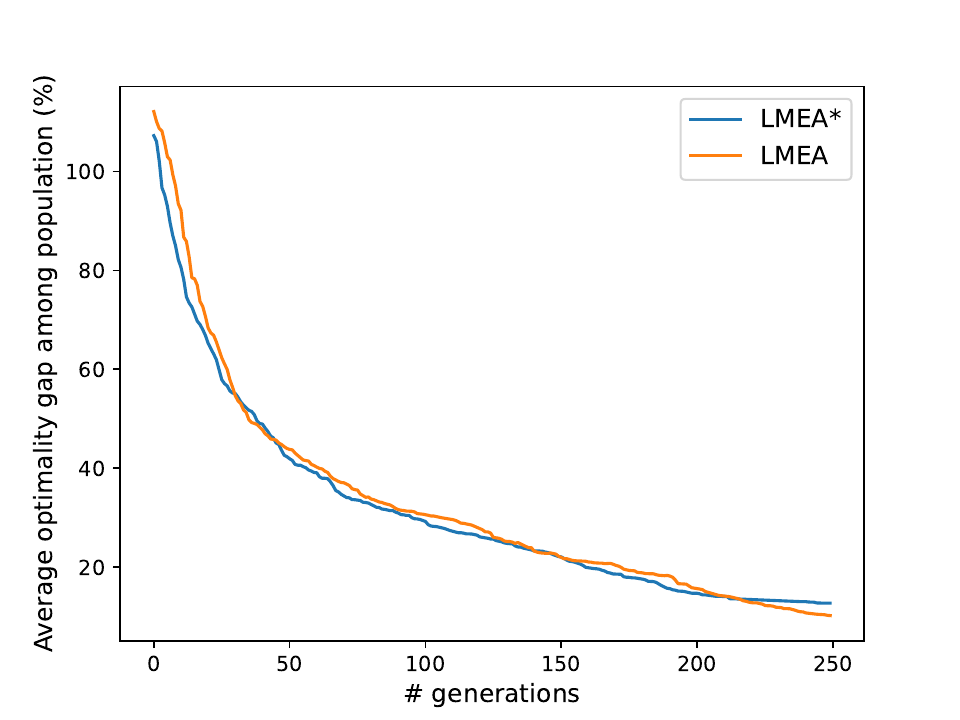}}
\caption{Convergence curves of LMEA and its variant without self-adaptation (LMEA*) on the rue-20 test set. 
	(a) Optimality gap of the best found solution as the generation number increases.
	(b) Average optimality gap of the population as the generation number increases.}
\label{fig:adaptive_20}
\end{figure}

\subsection{Effectiveness of Self-Adaptation}
To validate the effectiveness of the self-adaptation of the LLM's temperature in LMEA, we tested a variant of LMEA without self-adaptation (LMEA*) on the rue-20 test set.
The results are presented in Table~\ref{tab:self_adaptation}.
The convergence curves of LMEA and LMEA* are also illustrated in Figure~\ref{fig:adaptive_20}.
It can be clearly observed that, LMEA can achieve significantly better optimality gaps than LMEA*.
Moreover, based on Figure~\ref{fig:adaptive_20}, one can observe that even when the quality of the random initial solutions are worse than that of LMEA*, LMEA is still capable of finding better solutions more quickly.
These results demonstrate the effectiveness of self-adaptation.

\section{Discussions and Conclusion}
\label{sec:conclu}
In this work, we explored on employing LLMs as evolutionary combinatorial optimizers, where the LLM repeatedly generates offspring solutions based on the current population.
Our investigation demonstrates that LMEA has the capacity of solving non-trivial NP-hard combinatorial optimization problems such as TSPs.
Nonetheless, there are many open questions that remain to be explored in future works.
\begin{itemize}
		\item \textbf{Improving the scalability of LMEA.} Currently LMEA still has limitations in handling relatively large problems. One possible approach to improve the scalability of LMEA is to instruct it to only focus on improving the local parts of the solutions, rather than the whole solution.
	\item \textbf{Learning lessons from unsuccessful solutions.} Instructing LLMs to learn lessons from incorrect answers has been proven effective in improving their performance~\cite{shinn2023reflexion}. Hence, it is interesting to investigate how such a strategy can boost the performance of LMEA for solving optimization problems.
	\item \textbf{Reducing the runtime and cost of LMEA.} Currently, executing the full optimization process of LMEA on a small-scale problem is highly expensive and time-consuming (around half a day) due to frequent interactions with ChatGPT's API. 
	Future works can explore using smaller fine-tuned models for local execution to mitigate this issue.
	\item \textbf{Leveraging state-of-the-art prompt engineering.} Techniques like chain of thoughts~\cite{Wei0SBIXCLZ22} and self-consistency~\cite{Wang2023} have the potential to enhance the performance of LMEA.
	\item \textbf{Applying LMEA to other problems.} It is always interesting to investigate how LMEA would perform on different combinatorial optimization problems.
\end{itemize}

\section*{Acknowledgments}
This work is supported in part by the Centre for Frontier AI Research (CFAR), Agency for Science, Technology and Research (A*STAR), in part by the School of Computer Science and Engineering at Nanyang Technological University, and in part by the National Natural Science Foundation of China under Grant 62272210.
The research work is carried out in CFAR.


\bibliographystyle{IEEEtran}
\bibliography{IEEEabrv,mybib}

\end{document}